\documentclass[10pt,twocolumn,letterpaper]{article}

\usepackage{cvpr}      
\usepackage{float}
\usepackage{capt-of}

\definecolor{cvprblue}{rgb}{0.21,0.49,0.74}
\usepackage[pagebackref,breaklinks,colorlinks,allcolors=cvprblue]{hyperref}

\def\paperID{38831} 
\def\confName{CVPR}
\def\confYear{2026}

\title{Tri-Modal Fusion Transformers for UAV-based Object Detection}

\author{Craig Iaboni \\
New Jersey Institute of Technology\\
{\tt\small csi3@njit.edu}
\and
Pramod Abichandani\\
New Jersey Institute of Technology\\
{\tt\small pva23@njit.edu}
}

\begin{document}
\maketitle

\begin{figure*}[t!]
\centering
\includegraphics[scale=0.37]{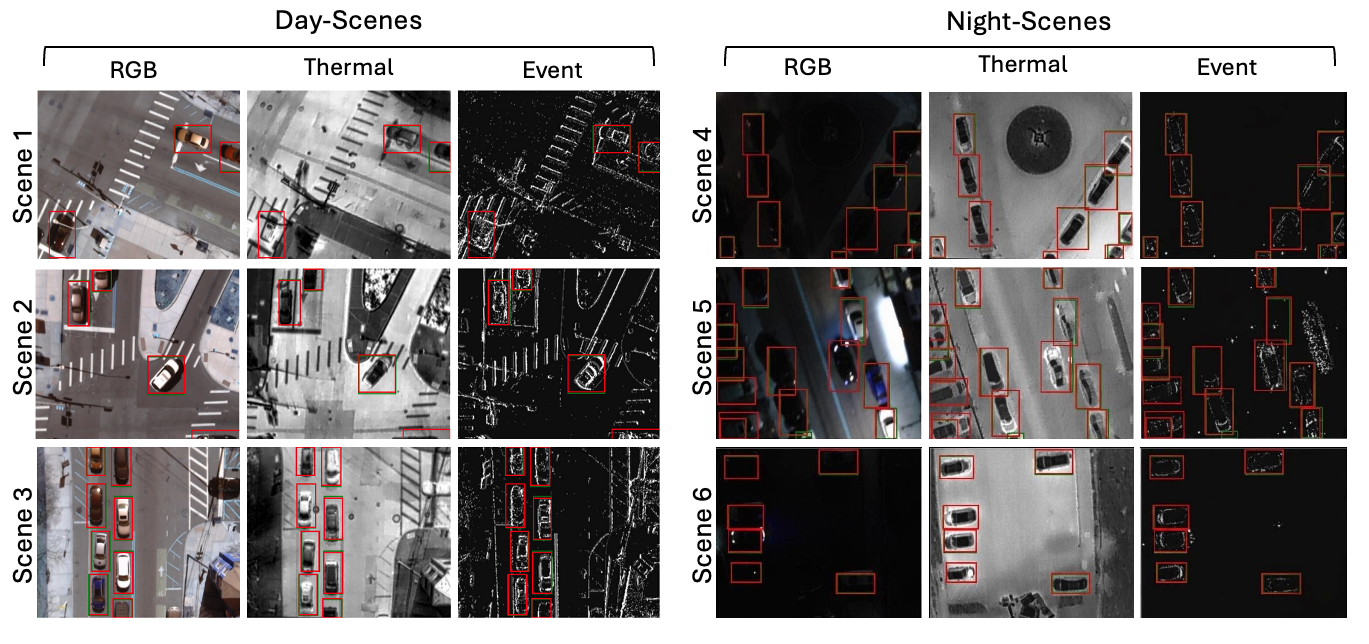}
\caption{Example tri-modal UAV dataset samples. 
Each scene shows synchronized RGB, thermal (LWIR), and event projections with annotated vehicles. Left: daytime samples illustrating illumination variation and viewpoint diversity. Right: nighttime samples highlighting the limitations of RGB and the complementary value of thermal and event cues for robust detection.}
\label{fig:dataset}
\end{figure*}

\begin{abstract}
Reliable UAV object detection requires robustness to illumination changes, motion blur, and scene dynamics that suppress RGB cues. Thermal long-wave infrared (LWIR) sensing preserves contrast in low light, and event cameras retain microsecond-level temporal edges, but integrating all three modalities in a unified detector has not been systematically studied. We present a tri-modal framework that processes RGB, thermal, and event data with a dual-stream hierarchical vision transformer. 
At selected encoder depths, a Modality-Aware Gated Exchange (MAGE) applies inter-sensor channel and spatial gating, and a Bidirectional Token Exchange (BiTE) module performs bidirectional token-level attention with depthwise–pointwise refinement, producing resolution-preserving fused maps for a standard feature pyramid and two-stage detector.

We introduce a 10{,}489-frame UAV dataset with synchronized and pre-aligned RGB–thermal–event streams and 24{,}223 annotated vehicles across day and night flights. Through 61 controlled ablations, we evaluate fusion placement, mechanism (baseline MAGE+BiTE, CSSA, GAFF), modality subsets, and backbone capacity. Tri-modal fusion improves over all dual-modal baselines, with fusion depth having a significant effect and a lightweight CSSA variant recovering most of the benefit at minimal cost. This work provides the first systematic benchmark and modular backbone for tri-modal UAV-based object detection. Code and dataset are available at \url{https://github.com/radlab-sketch/trimodal-uav-det}.

\end{abstract}

\vspace{-0.2in}
\section{Introduction}
UAV perception systems increasingly operate in conditions where no single sensor is reliable: visible-spectrum cameras lose discriminative structure under low light and motion, thermal sensors saturate or blur during rapid platform dynamics, and event cameras provide sparse but noisy evidence of motion \cite{gallego2020event}. Despite these complementary strengths and weaknesses, existing detection pipelines are overwhelmingly built around RGB or, at best, dual-modal pairings \cite{hou2024object}. This has left a fundamental architectural question unresolved: how should three heterogeneous sensing modalities interact inside a modern detector so that each compensates for the others’ failure modes? UAVs experience independent failure modes: illumination collapse, platform blur, fast scene motion, atmospheric effects, and thermal clutter \cite{sun2022drone}. RGB, thermal, and event sensors each address only a subset of these conditions, and no modality pair remains reliable across all of them \cite{nguyen2021review}. This motivates a detector that can selectively rely on the modality that remains informative as conditions vary within and across frames.

Tri-modal fusion is challenging for reasons that go beyond simply stacking channels. LWIR imagery reflects radiometric contrast rather than texture; event streams encode asynchronous temporal changes with no absolute intensities; and RGB provides high-resolution structure but collapses under illumination shifts \cite{mishra2025nystromformer, gehrig2018asynchronous}. These modalities differ in noise characteristics, spatial alignment sensitivity, temporal density, and semantic reliability across conditions. Conventional early-fusion approaches ignore these differences, while late-fusion pipelines forfeit the ability to shape intermediate representations jointly \cite{tziafas2023early}. Transformer backbones offer natural interfaces for cross-modal exchange, but how and at which resolutions fusion should occur has not been systematically explored. 

The present work treats tri-modal fusion as an architectural design space rather than a feature-level add-on. Our backbone maintains separate streams where doing so preserves modality-specific structure and couples them only at chosen intermediate stages, enabling controlled study of when, where, and how fusion is effective. 
This framework supports quantitative comparison of fusion operators, fusion depth, modality combinations, and backbone capacity, dimensions that prior work on multimodal detection has not evaluated under a single, controlled setting. Existing RGB-thermal and RGB–event datasets do not provide synchronized tri-modal frames or resolution-aligned annotations, making controlled tri-modal fusion studies impossible without constructing a dedicated dataset. 

\section{Related Work}
\label{sec:relatedworks}
{\bf Multi-Modal Fusion for Object Detection:} Single-modality detectors built on RGB imagery \cite{girshick2014rich, ren2015faster} degrade sharply under low illumination, motion blur, and adverse environmental conditions \cite{hwang2015multispectral}. This has motivated extensive work on pairing RGB with complementary sensing modalities. RGB–thermal methods \cite{sun2020fuseseg, zhang2021deep, yang2023comparison, choi2018kaist, liu2016multispectral, shivakumar2020pst900} leverage LWIR contrast to recover targets when visible cues collapse. RGB–event approaches \cite{zhou2023rgb, tomy2022fusing, lichtsteiner2008128, gehrig2019end} exploit microsecond-level temporal changes to stabilize detection under rapid motion. Thermal–event fusion has also been explored \cite{geng2024event}, using the temporal sparsity of events to refine low-light thermal signatures.

Despite extensive dual-modal work, nearly all prior efforts restrict fusion to \emph{two} modalities tailored to specific operating regimes (e.g., night-time RGB–thermal or high-speed RGB–event detection). 
To the best of our knowledge, no existing detector integrates RGB, thermal, and event sensing within a unified architecture, and there is no tri-modal UAV benchmark for systematically studying how the three modalities should interact. This paper addresses this gap by providing both a tri-modal dataset and a controlled fusion framework for evaluating modality combinations and fusion behavior under realistic UAV conditions. 

{\bf Fusion Mechanisms and Attention Operators:}
A core challenge in multimodal detection is determining where and how heterogeneous sensor streams should interact. Classical taxonomies distinguish early, late, and intermediate fusion \cite{baltruvsaitis2018multimodal}, but most multimodal detectors employ either simple channel concatenation at the input \cite{simonyan2014two} or high-level feature merging near the output \cite{karpathy2014large}. 
Intermediate fusion, where modalities exchange information within the backbone, has shown stronger performance \cite{feichtenhofer2016convolutional}, but the design space remains largely unexplored, especially for more than two modalities.

A wide range of operators have been proposed for cross-modal feature interaction. Channel gating mechanisms such as squeeze–excitation (SE) \cite{althoupety2024daff, zhao2025rethinking} and efficient channel attention (ECA) \cite{wang2020eca} modulate activations based on global responses, while spatial attention modules guide fusion to locations where modalities agree \cite{liang2023explicit}. Several architectures combine these ideas: GAFF \cite{zhang2021guided} uses spatial guidance masks to fuse RGB–thermal features; CGFNet \cite{wang2021cgfnet} alternates guided and cross-guided fusion blocks across scales; CMAFF \cite{wang2024cross} introduces uncertainty-aware weighting; and CSSA \cite{cao2023multimodal} replaces low-salience channels with their cross-modal counterparts before spatial selection. Other approaches perform iterative or multi-stage refinement, such as cyclic fuse-and-refine blocks \cite{zhang2020multispectral} or repeated cross-attention with weight sharing \cite{shen2024icafusion}. 

Most existing operators therefore address only RGB–thermal or RGB–event fusion, and little is known about their behavior when extended to tri-modal inputs or to hierarchical transformer backbones. In contrast, our study embeds multiple fusion families (including channel/spatial gating (GAFF, CSSA) and cross-attention-style interactions) into a unified transformer framework, enabling controlled comparisons of fusion depth and mechanism under identical architectural and detection settings. Unlike CSSA-style hard channel replacement or GAFF-style guided residual merging, our baseline block combines joint-conditioned cross-residual gating with token-level exchange while preserving stage stride and width, making it directly pluggable at different depths of the hierarchical backbone.

{\bf Hierarchical Backbones for Multimodal Vision:}
Multimodal detectors commonly use separate backbone streams because modality-specific low-level structure limits full parameter sharing \cite{cao2023multimodal, zhao2025rethinking}. Hierarchical transformers are particularly well suited to this setting because they provide multi-scale token representations and natural interfaces for intermediate cross-modal interaction \cite{dosovitskiy2020image, liu2021swin, chen2021crossvit}. Mix-Transformer (MiT) encoders \cite{xie2021segformer} are especially attractive for stage-wise fusion due to their overlapping patch embeddings and multi-resolution feature hierarchy, and have already been adopted in dual-encoder multimodal frameworks such as CMX \cite{zhang2023cmx}. For event-based vision, prior work has also emphasized that fusion mechanisms must respect the sparse and asynchronous statistics of event data \cite{gallego2020event}.

Existing multimodal backbones primarily target two-modality fusion and do not expose a systematic design space for studying fusion depth or fusion operators in the transformer hierarchy. In contrast, our framework adopts MiT-style hierarchical encoders for separate RGB and auxiliary streams and inserts fusion modules at explicit, resolution-aligned stages, enabling controlled comparison of multimodal backbones and fusion behavior under tri-modal inputs.

 \section{System Setup}

\begin{figure}[t!]
\centering
\includegraphics[scale=0.34]{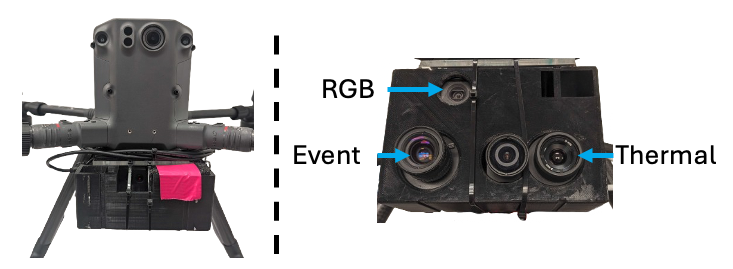}
\caption{Tri-modal UAV payload showing the underside-mounted sensor pod (left) and the co-aligned RGB, event, and thermal cameras in the custom housing (right).}
\label{fig:hardware}
\end{figure}

\subsection{Hardware}
\label{sec:hardware}
The tri-modal payload comprises three synchronized cameras mounted in a custom 3D-printed housing with fixed baselines and approximately parallel optical axes to ensure stable overlap across fields of view (Fig.~\ref{fig:hardware}). The RGB stream is captured by a Logitech HD camera at $1920{\times}1080$ and 30\,FPS. Thermal imagery is provided by a FLIR Duo operating in the 7.5–13.5\,$\mu$m LWIR band at $640{\times}512$ and 30\,FPS. Event data is recorded by a Prophesee VGA contrast-sensing sensor ($640{\times}480$ pixels, $15\,\mu\mathrm{m}$ pitch, $>120$\,dB dynamic range).

All three sensors are rigidly integrated in the housing and time-stamped by an on-board NVIDIA Jetson Xavier, which handles acquisition and storage. The payload is powered by a dedicated battery pack and flown on a DJI M300 platform~\cite{DJI_M300_2020}. The housing geometry is chosen to maintain balanced weight distribution and to avoid obstructing the M300 airframe and ventilation. Intrinsic and extrinsic calibration was performed once per flight session, and reprojection error was maintained below 1.5 pixels across modalities.

\subsection{Dataset}
We evaluate fusion strategies on a tri-modal dataset acquired with the platform in Sec.~\ref{sec:hardware}. Each sample consists of a pre-aligned, five-channel image tensor and YOLO-format labels. Arrays $\mathbf{X}\in\mathbb{R}^{H\times W\times 5}$ are stored in \texttt{.npy} format, with channels 0–2 as RGB, channel 3 as thermal, and channel 4 as an event frame; labels are per-image YOLO text files. The corpus contains 10{,}489 images with 24{,}223 bounding boxes for a single class (vehicle). We focus on a single vehicle class to establish a controlled tri-modal benchmark with high-quality cross-modal annotations. All modalities are pre-warped to a common plane and resolution, and most images are standardized to $301{\times}391$ pixels. Streams are synchronized via Jetson hardware timestamps. 

Data was collected over multiple UAV flights over an urban university campus under varied traffic and illumination conditions. The dataset spans varied time-of-day, with 6{,}412 day images and 4{,}077 night images (61.1\% day / 38.9\% night). The night portion is used to stress RGB-only detectors and to highlight the contribution of thermal and event cues.

A semi-automatic labeling protocol was used. For daytime sequences, a pretrained YOLO detector generated only initial candidate boxes on RGB frames. These candidates were projected into the common image plane using the cross-modal calibration and then exhaustively reviewed by human annotators, who added missed objects, removed false positives, and corrected box extents. For night-time sequences, RGB proposals were unreliable, so all images were labeled manually in the thermal plane before projection and then underwent a second-pass quality check. Final labels are therefore human-validated annotations in the shared image plane rather than direct outputs of the proposal model.

To assess proposal-seeding bias, we manually re-audited 723 randomly sampled frames by comparing the initial proposal set after the final human-validated annotations. Across these frames, 263 of 2411 final boxes (10.9\%) were absent from the initial proposals and were added by annotators. Of the 263 boxes recovered by annotators but absent from the initial proposal set, 168 were small or distant (\textless25 px), 63 were heavily occluded (\textless50\% visible), and 32 were truncated at image boundaries. These misses were therefore concentrated in difficult cases, indicating that the proposal stage served as an initialization aid while final labels were determined by manual verification and correction.

Event frames were generated by binning polarity events within a fixed temporal window $\Delta t \approx 33.3$\,ms (matching the 30\,FPS frame interval) centered at each RGB/thermal timestamp, followed by per-window normalization. Polarities were preserved, and events within each window were formed into an activation map where each pixel signaled the presence of ON/OFF events prior to normalization. Before training, each channel was affinely normalized: RGB channels used ImageNet statistics, while thermal and event channels used statistics computed from the training split. 

\vspace {-0.1in}
\section{Method}

\subsection{Model Architecture}

\label{sec:model-arch}

\begin{figure*}[t]
\centering
\includegraphics[scale=0.55]{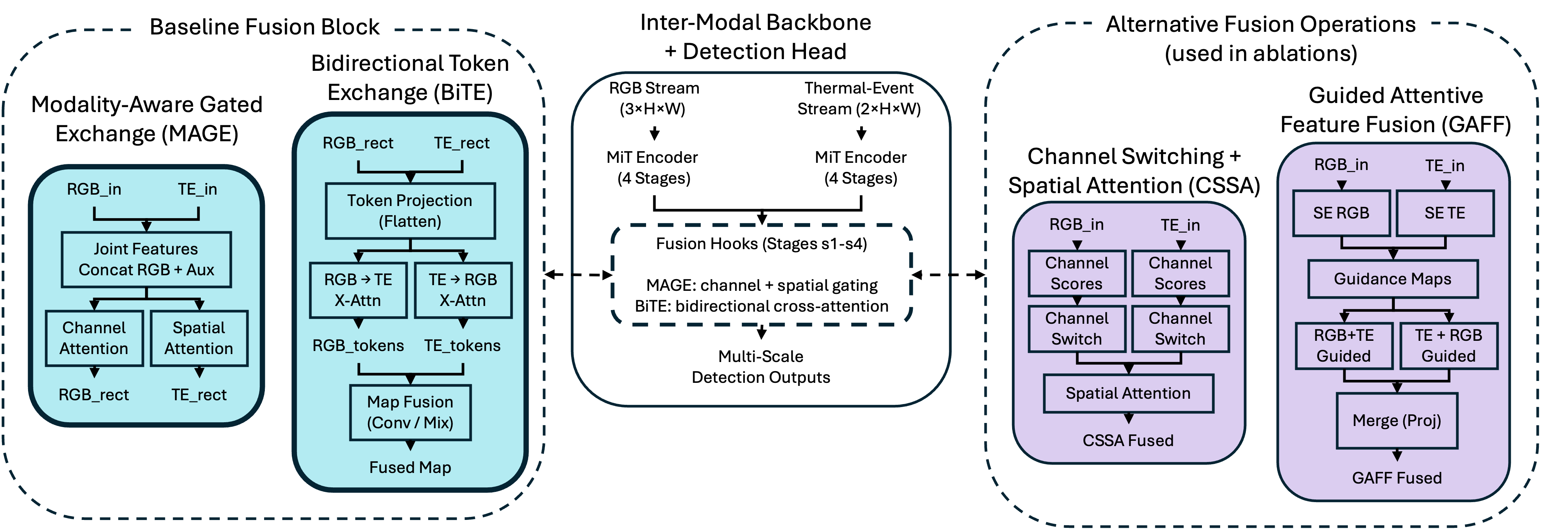}
\caption{Overview of the tri-modal detection framework.
Left: the baseline fusion block, combining Modality-Aware Gated Exchange (MAGE) and Bidirectional Token Exchange (BiTE).
Center: dual-stream MiT backbone with fusion hooks at stages s1–s4, producing multi-scale features for a two-stage detector.
Right: alternative fusion operators used in ablations: CSSA (channel switching + spatial gating) and GAFF (SE-based guided fusion).}
\label{fig:archi}
\end{figure*}
Our detector operates on tri-modal inputs and combines a dual-stream hierarchical transformer with stage-wise cross-modal interaction and a standard FPN-based two-stage detection head (Fig.~\ref{fig:archi}). Each sample is a five-channel tensor $X\in\mathbb{R}^{B\times 5\times H\times W}$ formed by stacking three RGB channels with one thermal and one event channel. The input is partitioned into an RGB stream $X_{\mathrm{rgb}}$ and a thermal–event (TE) stream $X_{\mathrm{TE}}$.

Both streams are processed by identical four-stage Mix-Transformer (MiT) backbones with independent weights. Stages perform overlapping patch embedding and multi-head self-attention with stage-dependent spatial reduction, producing multi-scale feature maps at strides $\{4,8,16,32\}$. At selected stages, the backbone inserts a fusion block that rectifies and merges the two streams while preserving spatial resolution and channel width. Fused outputs from all active stages are passed unchanged to a five-level FPN, which feeds a standard Faster~R-CNN head for region proposal and classification. Because the backbone maintains fixed shapes across fusion placements, the neck and detection head require no architectural changes for any of the ablations.
We group thermal and event channels into a single auxiliary stream to reduce backbone redundancy. Under the main tri-modal MiT-B1 setting, the default two-stream design (RGB vs.\ TE) achieves 84.24\% mAP with 60.01M parameters. A three-stream variant with separate RGB, thermal, and event encoders increases model size to 88.18M parameters without providing a meaningful accuracy gain in our preliminary comparisons, so we adopt the two-stream formulation as the default. This choice is also better aligned with UAV SWaP constraints, where additional backbone redundancy increases memory use, latency, and onboard power demand.

\subsection{Hierarchical Inter-Sensor Backbone}
Each stream is encoded by a four-stage hierarchical transformer following the MiT design. Stage~1 applies a $7{\times}7$/s4 overlapping patch embedding to the raw input; stages~2–4 use $3{\times}3$/s2 embeddings on the previous stage’s output. Tokens are processed by transformer blocks with pre-norm LayerNorm, spatial-reduction attention, and a feed-forward network containing a depthwise $3{\times}3$ convolution between the two linear layers, which restores local spatial coupling.

For a $224{\times}224$ reference input, the per-stage spatial resolutions are $56{\times}56$, $28{\times}28$, $14{\times}14$, and $7{\times}7$ with channel widths $\{64,128,320,512\}$. Both streams follow the same resolution schedule, ensuring shape alignment at every stage. At the end of each stage, tokens are reshaped back to feature maps and passed to the inter-sensor interaction module described in Sec.~\ref{sec:fusion-blocks}. Stages not selected for fusion simply propagate their modality-specific maps forward. This design exposes four resolution-aligned fusion hooks without altering the downstream detector interface, enabling controlled experiments on fusion depth and fusion mechanism.

\subsection{Stage-Wise Rectification and Fusion}
\label{sec:fusion-blocks}
Fusion occurs at a subset of the four backbone stages. At each selected stage, two submodules operate on the RGB and TE feature maps
$\mathbf{x}^{\mathrm{rgb}}, \mathbf{x}^{\mathrm{TE}} \in \mathbb{R}^{B\times C\times H\times W}$:
(1) a Modality-Aware Gated Exchange (MAGE) that performs cross-modal channel and spatial gating, and
(2) a Bidirectional Token Exchange (BiTE) that aggregates gated streams into a single fused representation while preserving spatial size and width.

\subsubsection{Modality-Aware Gated Exchange (MAGE)}
MAGE computes channel and spatial gates from the concatenated descriptor
\[
\mathbf{z} = [\mathbf{x}^{\mathrm{rgb}} \,\|\, \mathbf{x}^{\mathrm{TE}}] \in \mathbb{R}^{B\times 2C\times H\times W},
\]
allowing each stream to be modulated based on joint evidence from both modalities rather than on single-stream statistics.

{\bf Channel gating}: Global average and max pooling of $\mathbf{z}$ form complementary global summaries, which are passed through a two-layer $1{\times}1$ MLP (nonlinearity + sigmoid) to produce directional per-channel gates
\[
w^{c}_{\mathrm{TE}\rightarrow \mathrm{rgb}},\;
w^{c}_{\mathrm{rgb}\rightarrow \mathrm{TE}}
\in [0,1]^{B\times C\times 1\times 1}.
\]
Gates modulate only the cross-stream residuals, leaving each stream’s identity path unchanged. This preserves modality-specific structure while allowing cross-modal reinforcement of channels that are consistently informative.

{\bf Spatial gating}:
A lightweight $1{\times}1 \!\rightarrow$ nonlinearity $\!\rightarrow 1{\times}1$ head predicts pixelwise masks
\[
w^{s}_{\mathrm{TE}\rightarrow \mathrm{rgb}},\;
w^{s}_{\mathrm{rgb}\rightarrow \mathrm{TE}}
\in [0,1]^{B\times 1\times H\times W}
\]
from $\mathbf{z}$. These masks scale only the cross-residual updates, restricting spatial transfer to locations where modalities exhibit consistent evidence and suppressing transfer in noisy or modality-specific regions. The outputs of MAGE are the rectified feature maps
\begin{align}
\hat{\mathbf{x}}^{\mathrm{rgb}}
&= \mathbf{x}^{\mathrm{rgb}}
\;+\;
w^{s}_{\mathrm{TE}\rightarrow\mathrm{rgb}}
\cdot
\big(
  w^{c}_{\mathrm{TE}\rightarrow\mathrm{rgb}}
  \cdot
  \mathbf{x}^{\mathrm{TE}}
\big),
\\[4pt]
\hat{\mathbf{x}}^{\mathrm{TE}}
&= \mathbf{x}^{\mathrm{TE}}
\;+\;
w^{s}_{\mathrm{rgb}\rightarrow\mathrm{TE}}
\cdot
\big(
  w^{c}_{\mathrm{rgb}\rightarrow\mathrm{TE}}
  \cdot
  \mathbf{x}^{\mathrm{rgb}}
\big).
\end{align}
    

\subsubsection{Bidirectional Token Exchange (BiTE)}
BiTE fuses the rectified maps by symmetric cross-attention and lightweight spatial refinement. Flattening
$\hat{\mathbf{x}}^{\mathrm{rgb}}, \hat{\mathbf{x}}^{\mathrm{TE}}$ into token sequences
$\mathbf{T}_{\mathrm{rgb}}, \mathbf{T}_{\mathrm{TE}}\in\mathbb{R}^{B\times N\times C}$ ($N=H\!W$), we form projections
\begin{equation}
\begin{aligned}
\mathbf{Q}_s &= \mathbf{T}_s \mathbf{W}_s^{Q}, \qquad
\mathbf{K}_{\bar{s}} = \mathbf{T}_{\bar{s}} \mathbf{W}_{\bar{s}}^{K},
\\
\mathbf{V}_{\bar{s}} &= \mathbf{T}_{\bar{s}} \mathbf{W}_{\bar{s}}^{V},
\qquad s \in \{\mathrm{rgb},\mathrm{TE}\}.
\end{aligned}
\end{equation}

and update each stream via cross-attention
\[
\tilde{\mathbf{T}}_s
= \mathbf{T}_s
+ \mathrm{Softmax}\!\left(\frac{\mathbf{Q}_s \mathbf{K}_{\bar{s}}^\top}{\sqrt{d_k}}\right)
  \mathbf{V}_{\bar{s}}.
\]
The updates are residual, preserving modality-specific content while introducing cross-modal context.

Concatenating the updated tokens,

\[
\mathbf{Z} = [\,\tilde{\mathbf{T}}_{\mathrm{rgb}}\,;\,\tilde{\mathbf{T}}_{\mathrm{TE}}\,]
\in \mathbb{R}^{B\times N\times 2C},
\]
and reshaping to maps yields $\mathbf{U}_0\in\mathbb{R}^{B\times 2C\times H\times W}$. A depthwise $3{\times}3$ convolution restores locality, and a $1{\times}1$ projection mixes channels and compresses the width back to $C$, producing the fused map
\[
\mathbf{u} \in \mathbb{R}^{B\times C\times H\times W}.
\]

BiTE preserves the spatial stride and channel width of the current stage, enabling flexible fusion placement and allowing the downstream FPN and detection head to remain unchanged.

\subsection{Fusion Placement}
Fusion hooks were inserted at the four resolution-aligned backbone stages (strides $\{4,8,16,32\}$). At each selected stage, the backbone replaces the two modality-specific outputs with a single fused map produced by MAGE and BiTE, while unselected stages forward both streams independently. Because fusion preserves spatial stride and channel width, all configurations (single-stage, multi-stage, or full-stage fusion) produce identical interfaces to the FPN and two-stage detector. 

\subsection{Feature Pyramid Neck}

We use a standard top-down Feature Pyramid Network that projects stage outputs to 256 channels and produces a five-level pyramid at strides $\{4,8,16,32,64\}$. The neck is fixed across all fusion configurations.

\subsection{Two-Stage Detection Head}

Detection is performed with a standard Faster R-CNN head over the five FPN levels, including an RPN, RoIAlign, and box/class prediction branches. Anchors, proposal assignment, loss functions, and head settings are held fixed across all experiments so that performance differences reflect backbone fusion behavior rather than detector changes.

\subsection{Ablation Protocol}
Fusion is treated as a pluggable operator at any of the four backbone stages. For a given configuration, we activate fusion at a chosen subset of stages while keeping all downstream components (FPN, RPN, RoI head, losses, and training schedule) identical. Because fusion preserves the spatial stride and channel width of each stage, all configurations supply the same feature shapes to the detector. This setup isolates two variables: (1) which stages perform fusion, and (2) which fusion mechanism is used at those stages. We evaluate single-stage, multi-stage, and full-stage fusion, as well as dual-modal and tri-modal variants, under identical data and optimization settings.

\subsubsection{CSSA: Channel Switching and Spatial Attention}
CSSA provides a lightweight alternative to the baseline fusion block. It first scores channels in each stream using global average pooling followed by a short 1D convolution and sigmoid activation. Channels with scores below a threshold $\tau$ are replaced by their same-index counterparts from the other stream, producing two switched tensors. A spatial gate is then predicted from the concatenated switched tensors via a small convolutional head and selects between them at each pixel. CSSA preserves spatial resolution and channel width, introduces only minor computational overhead, and can be inserted at any fusion hook. In our comparisons, CSSA replaces the baseline fusion block at the specified stages while leaving the rest of the architecture unchanged.

\subsubsection{GAFF: Guided Attentive Feature Fusion}
GAFF is a higher-capacity alternative that combines channel recalibration and spatially guided cross-modal updates. Each stream is first processed by a squeeze–excitation block to emphasize informative channels. Directional guidance maps are then predicted so each modality receives location-aware corrections from the other via residual injection. The resulting pair of feature maps is merged through either a direct $1{\times}1$ projection or a bottlenecked variant, producing a single fused map with the same spatial size and channel width as the backbone stage. As with CSSA, GAFF can be inserted at any fusion hook; stages without GAFF simply forward both modality-specific streams. Our evaluations compare GAFF and CSSA under fixed training, backbone, and detection settings.

\vspace {-0.1in}

\section{Experimental Evaluation and Results}
\label{sec:results}

\subsection{Training Setup}
All models are trained for 15 epochs with stochastic gradient descent (momentum 0.9, weight decay $1\times10^{-4}$) and a cosine learning-rate schedule with linear warm-up over the first 500 iterations. The base learning rate is 0.02 for a global batch size of 16 and scales linearly with batch size. Inputs use the pre-aligned native resolution ($301{\times}391$) and are padded to the nearest multiple of 32 on each side for FPN compatibility. Anchor settings follow Torchvision defaults (per-level sizes $\{32,64,128,256,512\}$ and aspect ratios $\{0.5,1.0,2.0\}$), and the RoI head uses RoIAlign with $7{\times}7$ pooled features. Longer schedules did not yield improvements and occasionally led to mild overfitting, consistent with the dataset size and MiT backbone capacity.

Experiments are implemented in PyTorch with Torchvision detection utilities, using mixed-precision training. Unless stated otherwise, all models share the same optimizer, schedule, data augmentations, and evaluation protocol. In total, 61 experiments are run, covering backbone capacity, modality combinations, GAFF and CSSA fusion variants, external RGB–thermal baselines, cross-dataset transfers, and day/night training splits.

\vspace {-0.05in}

\subsection{Backbone Capacity}
The impact of encoder size under the standard MAGE+BiTE baseline with tri-modal (RGB+Thermal+Event) inputs are reported in  Table~\ref{tab:backbone} via COCO mAP, mAP$_{50}$, and parameter count for MiT backbones B0–B4. Performance is non-monotonic. MiT-B1 attains the best mAP (84.24\%) with roughly one-third the parameters of B4. Larger backbones (B2–B4) do not translate additional capacity into improved detection, and B4 even drops below B0. MiT-B1 therefore offers the best accuracy–efficiency trade-off and is used as the default backbone in subsequent studies, while MiT-B0 provides a strong low-capacity option (80.63\% mAP with only 27.79M parameters). These non-monotonic trends are consistent with overfitting behavior on modest-sized detection datasets, where backbones larger than B1 exceed what modest-sized detection datasets can support under the fixed training schedule and therefore fail to convert additional capacity into better generalization.

\begin{table}[t]
\centering
\small
\begin{tabular}{|l|r|r|r|}
\hline
\textbf{Backbone} & \textbf{Params (M)} & \textbf{mAP} & \textbf{mAP$_{50}$} \\
\hline
MiT-B0 & 27.79  & 80.63 & 97.85 \\
MiT-B1 & 60.01  & \textbf{84.24} & 98.95 \\
MiT-B2 & 82.10  & 82.91 & 98.06 \\
MiT-B3 & 155.40 & 82.43 & 98.06 \\
MiT-B4 & 196.60 & 79.97 & 97.93 \\
\hline
\end{tabular}
\caption{Backbone capacity study (MAGE+BiTE, RGB+Thermal+Event).}
\label{tab:backbone}
\end{table}

\vspace {-0.05in}

\subsection{GAFF Ablations}
We next evaluate GAFF as a higher-capacity alternative to the baseline fusion. All GAFF runs replace MAGE+BiTE at the specified stages, while keeping the backbone, neck, head, and training schedule identical.

{\bf Placement}: 
Table~\ref{tab:gaff-p1-main} reports the Phase~1 placement study using a default GAFF configuration (SE ratio $r{=}4$, separate inter-modality guidance, direct merge). GAFF performs best when inserted once at deeper stages, with s4 highest and multi-stage insertion consistently weaker.

\begin{table}[t]
\centering
\small
\begin{tabular}{|l|l|r|r|}
\hline
\textbf{Stages} & \textbf{Mechanism} & \textbf{mAP} & \textbf{mAP$_{50}$} \\
\hline
s1      & GAFF (r=4, sep., direct) & 82.74 & 98.28 \\
s2      & GAFF (r=4, sep., direct) & 81.87 & 98.18 \\
s3      & GAFF (r=4, sep., direct) & 83.20 & 98.19 \\
s4      & GAFF (r=4, sep., direct) & \textbf{83.41} & 98.22 \\
s23     & GAFF (r=4, sep., direct) & 82.20 & 97.36 \\
s34     & GAFF (r=4, sep., direct) & 82.73 & 98.19 \\
s234    & GAFF (r=4, sep., direct) & 82.89 & 98.22 \\
s1234   & GAFF (r=4, sep., direct) & 81.93 & 98.23 \\
\hline
\end{tabular}
\caption{GAFF Phase~1: fusion placement with default hyperparameters.}
\label{tab:gaff-p1-main}
\end{table}

{\bf Mechanism variants}: 
Phase~2 sweeps GAFF’s internal hyperparameters at the best-performing depths (primarily s4 and s3): the SE reduction ratio ($r\in\{4,8\}$), shared vs.\ separate guidance weights, and direct vs.\ bottleneck merge. Results are summarized in Table~\ref{tab:gaff-p2}. At s4, the best configuration uses $r{=}8$, separate guidance, and a direct merge (83.78\% mAP). At s3, the strongest variant (84.02\% mAP) uses $r{=}4$, shared guidance, and a bottleneck merge. Across all settings, GAFF matches or slightly trails the baseline MAGE+BiTE fusion while offering an alternative design point that emphasizes SE-style recalibration and guided spatial updates.

\begin{table}[b!]
\centering
\small
\begin{tabular}{|l|r|l|l|r|r|}
\hline
\textbf{Stage} & \textbf{SE$_r$} & \textbf{Inter} & \textbf{Merge} & \textbf{mAP} & \textbf{mAP$_{50}$} \\
\hline
s4 & 4 & separate & bottleneck  & 82.77 & 97.46 \\
s4 & 4 & shared   & direct      & 82.15 & 98.20 \\
s4 & 4 & shared   & bottleneck  & 81.65 & 97.43 \\
s4 & 8 & separate & direct      & \textbf{83.78} & 98.24 \\
s4 & 8 & separate & bottleneck  & 82.53 & 97.41 \\
s4 & 8 & shared   & direct      & 83.20 & 98.31 \\
s4 & 8 & shared   & bottleneck  & 83.36 & 97.46 \\
s3 & 4 & separate & bottleneck  & 83.38 & 98.14 \\
s3 & 4 & shared   & direct      & 82.46 & 98.27 \\
s3 & 4 & shared   & bottleneck  & 84.02 & 98.25 \\
s3 & 8 & separate & direct      & 83.75 & 98.16 \\
\hline
\end{tabular}
\caption{GAFF Phase~2: mechanism variants at s3 and s4.}
\label{tab:gaff-p2}
\end{table}

\vspace {-0.1in}

\subsection{CSSA Ablations}
We then evaluate CSSA as a lightweight drop-in replacement for MAGE+BiTE. CSSA is inserted at single or combined stages (s1, s2, s3, s4, s23, s34, s1234) with thresholds $\tau \in \{0.3, 0.5, 0.7\}$ using a MiT-B1 backbone (Table~\ref{tab:cssa}).

CSSA shows a preference for early fusion. The best configuration applies CSSA only at stage~1 with $\tau{=}0.5$, reaching 83.44\% mAP. At the same threshold, deeper single-stage placements are slightly weaker (s2: 82.58\%, s3: 82.80\%, s4: 83.20\% mAP), and multi-stage variants never outperform their single-stage counterparts (e.g., s23: 82.32\%, s34: 81.66\%, s1234: 80.91\% mAP). Repeated channel switching across scales appears to erode modality-specific structure rather than enhancing it. 

Varying $\tau$ primarily fine-tunes performance. At s1, mAP varies modestly (83.23\%, 83.44\%, 82.98\% for $\tau{=}0.3,0.5,0.7$), indicating robustness to threshold choice, with $\tau{=}0.5$ providing the best balance between aggressive channel replacement and redundancy preservation. Similar but smaller trends appear at deeper stages.


\begin{table}[t]
\centering
\small
\setlength{\tabcolsep}{4pt}
\begin{tabular}{|l|ccc|}
\hline
\textbf{Stages} & \multicolumn{3}{c|}{\textbf{Threshold $\tau$ (mAP / mAP$_{50}$)}} \\
\cline{2-4}
{} & 0.3 & 0.5 & 0.7 \\
\hline
s1    & 83.23 / 97.46 & \textbf{83.44} / 97.45          & 82.98 / 97.41 \\
s2    & 82.12 / 97.33 & 82.58 / 97.42          & 82.04 / 97.27 \\
s3    & 79.83 / 97.14 & 82.80 / 97.34          & 79.52 / 97.27 \\
s4    & 83.09 / 98.19 & 83.20 / 98.28          & 82.73 / 98.10 \\
s23   & 81.70 / 97.37 & 82.32 / 98.20          & 81.44 / 97.22 \\
s34   & 81.12 / 97.37 & 81.66 / 98.22          & 80.97 / 97.15 \\
s1234 & 80.44 / 97.10 & 80.91 / 98.03          & 80.26 / 97.06 \\
\hline
\end{tabular}
\caption{CSSA ablation over fusion depth and channel-switch threshold. Each entry reports mAP / mAP$_{50}$.}
\vspace {-0.15in}
\label{tab:cssa}
\end{table}

\vspace {-0.05in}

\subsection{Modality Ablations and External Baselines}
To quantify the contribution of each sensor, we train three dual-modal variants with the same MiT-B1 backbone and MAGE+BiTE fusion using RGB+Thermal, Thermal+Event, and RGB+Event inputs (Table~\ref{tab:modality}). We also benchmark two external RGB–thermal detectors: YOLOv11-RGBT~\cite{wan2025yolov11} and DetFusion~\cite{sun2022detfusion} and evaluate cross-dataset performance on M3FD~\cite{liu2022target} and RTDOD~\cite{feng2023rtdod}.

\begin{table}[b!]
\centering
\small
\setlength{\tabcolsep}{4pt}
\begin{tabular}{|l|r|r|}
\toprule
\textbf{Configuration} & \textbf{mAP} & \textbf{mAP$_{50}$} \\
\midrule
\multicolumn{3}{l}{\textbf{Tri-/Dual-Modal (Ours, MiT-B1)}} \\
RGB+Thermal      & \textbf{83.42} & 98.22 \\
Thermal+Event    & 74.86          & 96.95 \\
RGB+Event        & 66.32          & 94.46 \\
\midrule
\multicolumn{3}{l}{\textbf{External RGB--Thermal}} \\
YOLOv11-RGBT     & 82.08          & --    \\
DetFusion        & 78.00          & --    \\
\midrule
\multicolumn{3}{l}{\textbf{Cross-Dataset (RGB--Thermal, Ours)}} \\
M3FD             & 81.79          & 97.36 \\
RTDOD            & 69.21          & 93.87 \\
\bottomrule
\end{tabular}
\caption{Modality ablations and RGB--thermal baselines. All ``Ours'' rows use the MiT-B1 backbone with MAGE+BiTE fusion. External models operate on RGB+Thermal inputs only.}
\label{tab:modality}
\end{table}

\begin{figure}[t]
\centering
\includegraphics[width=\columnwidth]{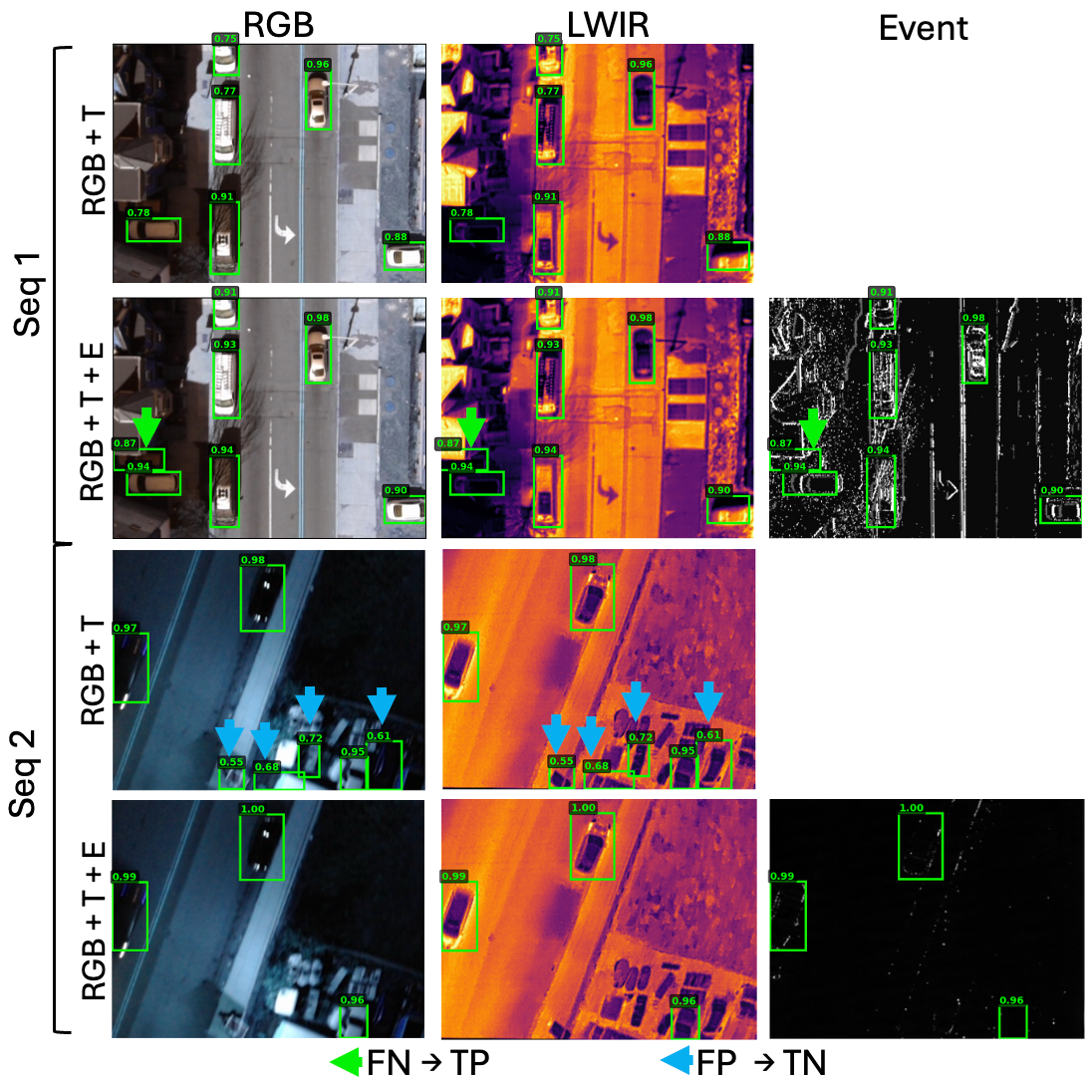}
\caption{Representative cases where adding events improves detection relative to RGB+Thermal under the same detector and training protocol. Left: false-negative recovery in motion-affected or visually ambiguous regions. Right: false-positive suppression at night when thermal clutter is not supported by coherent event evidence.}
\label{fig:event-help-panel}
\end{figure}

Among dual-modal variants, RGB+Thermal is the strongest pair (83.42\% mAP), while the best tri-modal setting yields a modest additional gain, indicating that most of the benefit is captured by combining RGB with thermal and that events contribute primarily in specific failure regimes. Thermal+Event (74.86\% mAP) is substantially weaker, and RGB+Event performs worst (66.32\% mAP), consistent with thermal being the most informative secondary modality for UAV detection. 
Figure~\ref{fig:event-help-panel} illustrates two representative automatically mined cases where events provide complementary value: false-negative recovery in motion-affected or visually ambiguous regions, and false-positive suppression in night-time thermal clutter.

YOLOv11-RGBT reaches 82.08\% mAP on the proposed UAV dataset, slightly below the MiT-B1 tri-modal baseline, while DetFusion attains 78.00\%. On M3FD and RTDOD, the MiT-B1 RGB+Thermal model with MAGE+BiTE transfers reasonably to established RGB-thermal benchmarks. External baselines were trained from public implementations using the authors' recommended settings, adapted only for image resolution and normalization.


\begin{table}[t]
\centering
\small
\begin{tabular}{|l|r|}
\hline
\textbf{Fusion variant} & \textbf{mAP} \\
\hline
BiTE-only         & 76.88 \\
MAGE-only         & 81.01 \\
MAGE+BiTE         & 84.24 \\
\hline
\end{tabular}
\caption{Ablation isolating the two components of the baseline fusion block under the main MiT-B1 tri-modal setting.}
\label{tab:mage-bite}
\end{table}

To isolate the contributions of the two components in the baseline fusion block, we evaluated BiTE-only and MAGE-only variants under the main MiT-B1 tri-modal setting. BiTE-only removes reliability weighting and applies token exchange directly to the two streams, whereas MAGE-only removes token exchange and instead applies a minimal $2C \rightarrow C$ merge after gated rectification. Both alternatives underperform the full MAGE+BiTE design.

\vspace {-0.1in}

\subsection{Daytime vs.\ Night-time Training}
Finally, we quantify the effect of illumination diversity in training. Three MiT-B1 tri-modal models are trained using only daytime images, only night-time images, or the full day+night training set, keeping all other settings fixed (Table~\ref{tab:daynight}).
Models trained on a single illumination regime overfit to that regime: the day-only model fails to generalize to night scenes, and the night-only model underperforms on daytime images. Training on the full day+night corpus yields the best overall performance and a more balanced trade-off between daytime and night-time accuracy, underscoring the importance of diverse illumination for tri-modal UAV detection. 

\vspace {-0.1in}

\section{Conclusion}
We presented a tri-modal RGB–thermal–event detection framework and a synchronized UAV dataset that enable controlled evaluation of cross-modal fusion. Systematic ablations show that tri-modality consistently outperforms dual-modality, and that fusion depth and mechanism are decisive factors, with lightweight CSSA effective at early stages and higher-capacity GAFF most useful when applied selectively at deeper layers. The framework and dataset establish a reproducible basis for future work on temporal tri-modal fusion and adaptive modality selection.

\begin{table}[t!]
\centering
\small
\setlength{\tabcolsep}{4pt}
\begin{tabular}{|l|r|r|r|}
\toprule
\textbf{Split} & \textbf{All} & \textbf{Day} & \textbf{Night} \\
\midrule
Day-only        & 79.0  & 85.0 & 70.5 \\
Night-only      & 77.5  & 72.0 & 84.5 \\
Full day+night  & 82.24 & 84.0 & 80.0 \\
\bottomrule
\end{tabular}
\caption{Impact of illumination diversity on tri-modal detection.}
\label{tab:daynight}
\end{table}

{
    \small
    \bibliographystyle{ieeenat_fullname}
    \bibliography{refs}
}


\end{document}